\pdfoutput=1
\documentclass[11pt]{article}
\usepackage[final]{acl}

\usepackage[T1]{fontenc}
\usepackage[utf8]{inputenc}
\usepackage{newtxtext}
\usepackage{newtxmath}
\usepackage{inconsolata}
\usepackage{microtype}

\usepackage{siunitx}
\usepackage{graphicx}
\usepackage{tabularx}
\usepackage{booktabs}
\usepackage{hyperref}
\usepackage{balance}
\title{%
    Reference Games as a Testbed for the\\ 
    Alignment of Model Uncertainty and Clarification Requests
}

\author{%
    Manar Ali\textsuperscript{1,2} \:
    Judith Sieker\textsuperscript{3} \:
    Sina Zarrieß\textsuperscript{1,3} \:
    Hendrik Buschmeier\textsuperscript{1,2}
\\
\textsuperscript{1}%
         CRC 1646 ‘Linguistic Creativity in Communication’, Bielefeld University, Germany\\
     \textsuperscript{2}%
        Digital Linguistics Lab, Bielefeld University, Germany\\ 
    \textsuperscript{3}%
        Computational Linguistics Group, Bielefeld University, Germany\\
        \texttt{\{manar.ali|j.sieker|sina.zarriess|hbuschme\}@uni-bielefeld.de}\\
    }

\begin{document}
\maketitle

\begin{abstract}
    In human conversation, both interlocutors play an active role in maintaining mutual understanding. When listeners are uncertain about what speakers mean, for example, they can request clarification. It is an open question for language models whether they can assume a similar listener role, recognizing and expressing their own uncertainty through clarification. We argue that reference games are a suitable testbed to approach this question as they are controlled, self-contained, and make clarification needs explicit and measurable. To test this, we evaluate three vision-language models comparing a baseline reference resolution task to an experiment where the models are instructed to request clarification when uncertain. The results suggest that even in such simple tasks, models often struggle to recognize internal uncertainty and translate it into adequate clarification behavior. This demonstrates the value of reference games as testbeds for interaction qualities of (vision and) language models.\footnote{%
        Code and data are available as a persistent data publication at \url{https://doi.org/10.5281/zenodo.20158390} and, additionally, for your convenience, at \url{https://github.com/Manarali-bit/reference-games-clarification}.}
\end{abstract}

\section{Introduction}
\label{sec:introduction}

Ambiguity and misunderstanding in dialogue is inevitable \citep{weigand1999misunderstanding}. Problems in understanding and confusion can arise even for simple utterances, since speakers and addresses process speech subjectively and audience design in language production can never be perfect.
For example, a request like “Can you pass me the blue pencil?” may prompt uncertainty in a situation where several blueish pencils are available, as color perception is subjective \citep{bimler2004quantifying}
and context-specific \citep{mitterer2008recalibrating}.
However, if a listener cannot resolve an ambiguity, communication need not fail. They can initiate \emph{repair}, e.g., by asking a \emph{clarification request} (i.e., “The light blue one?”), which explicitly signals uncertainty and requests additional information. Repair and clarification are fundamental mechanisms for achieving intersubjectivity and mutual understanding, employed frequently by humans \citep[on average every 84 seconds, as found crosslinguistically by][]{dingemanse2015universal}. 
Often, it is the listener who takes the lead in signaling an ambiguity, making mutual understanding a shared responsibility \citep{Clark_WilkesGibbs_1986, Clark_1996, dingemanse2023interactive}. 

For language models, however, it is still an open question whether they employ similar mechanisms when facing uncertainty in dealing with human input. Although they can generate fluent and contextually appropriate responses, this very fluency can mask underlying comprehension problems, leading humans to overestimate model competence \citep{sieker-etal-2024-illusion, rathi2025} and undermining trust in collaboration \citep{dhuliawala-etal-2023-diachronic, si-etal-2024-large}.
At the same time, their linguistically communicated confidence is often poorly calibrated to their actual accuracy \citep{MielkeSzlam2022}, raising the question of whether they can act appropriately given uncertainty, e.g., by requesting clarification -- an ability crucial for preventing misunderstandings from being hidden behind confident output.

Addressing the question of whether models can signal their uncertainty through clarification, however, is difficult in open dialogue settings, where there is no ground truth for when clarification is warranted and the space of possible interpretations is open-ended. Here, we propose \emph{reference games} \citep{Clark_WilkesGibbs_1986, frank2012predicting} as a testbed that makes clarification needs explicit and measurable. In their basic form, they involve two participants: a speaker and a listener, who share a set of candidate referents (e.g., images or abstract illustrations). The speaker's task is to describe a target object, and the listener must identify it among the set of candidates. Unlike in broader dialogue research, where clarification tends to focus on identifying a user’s underlying need or intent, reference games are goal-directed with fixed alternatives and require no external knowledge: it is immediately clear when a description fails to single out the target and a clarification is needed.

In this paper, we argue that reference games offer a controlled way to test whether language models can convert internal uncertainty into appropriate clarification requests. We investigate whether vision-language models (VLMs), as listeners in a reference game, are able to recognize their own uncertainty and respond with clarification questions.

\section{Background} 
\label{sec:background}

While research has made significant progress toward modeling dialogue behavior, including clarification and repair, recent analyses show that LLMs still fall short in core interactional mechanisms such as turn-taking, feedback, and repair \citep{PilanPrevot2024, StokoeAlbert2024, Mayor_etal_2025}. Furthermore, LLMs are far less likely than humans to engage in grounding acts, such as rejecting problematic input \citep{lachenmaier-etal-2025-llms, Sieker2025}, often presuming common ground instead \citep{shaikh-etal-2024-grounding}.
Crucially, LLMs rarely initiate repair proactively, typically relying on humans to do so \citep{Puetz_2023_ChatGPTrepair}. When they clarify, their clarification requests are often miscalibrated compared to human patterns \citep{madge2025referentialambiguityclarificationrequests}. 

This raises the central question of whether language models can appropriately act on their own uncertainty by formulating clarification requests. Quantifying uncertainty has become an increasingly central topic in NLP and the study of large language models \citep{VashurinFadeeva2025, Shorinwa_2025_uncertaintyquantification},
yet few studies have connected these uncertainty quantification to clarification behavior.
\citet{testoni-fernandez-2024-asking} and \citet{zhang_clarify_2025}, for example, propose entropy-based uncertainty measures to determine whether a clarification should be requested. However, such approaches typically focus on open-ended intent understanding tasks, where the space of possible interpretations is difficult to delimit. Here, reference games offer a controlled alternative. Reference games have long served as controlled settings in (computational) linguistics to study how interlocutors establish shared understanding \citep{Clark_WilkesGibbs_1986, brennan1996conceptual}, and to evaluate model competence in resolving referring expressions \citep{monroe-etal-2017-colors}.
Recent work has shown that even state-of-the-art VLMs struggle in these tasks, particularly when distractors are similar, the input has nested structure, or descriptions are underspecified \citep{junker-etal-2025-multimodal, testoni2025racquet}. These findings suggest that reference games remain a useful benchmark for probing models’ pragmatic competence. 
\citet{khalid2020combining}, for instance, used them to study clarification behavior directly, combining reinforcement learning with cognitive user modeling to learn strategies depending on the structure of ambiguity in the interaction.

In this paper, we build on this line of work and show that reference games can be used not only to evaluate reference resolution, but more broadly, to test whether VLMs can express their internal uncertainty through clarification behavior.

\section{Methodology}
\label{sec:methodology}

\subsection{Models and Data}
\label{sec:models-data}

\begin{figure}
    \centering
    \includegraphics[width=\columnwidth]{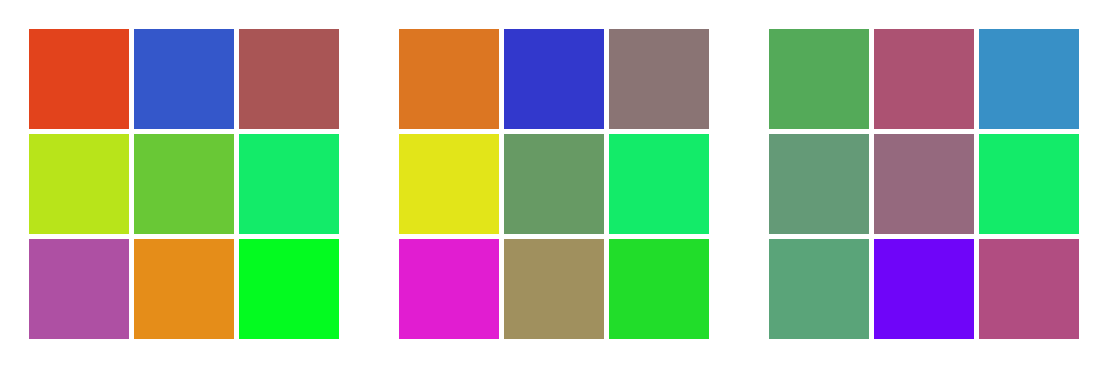}
    \caption{Example item from the dataset \citep{mcdowell-goodman-2019-learning}. The speaker referred to the second item with the description “Bottom left is bright pink.”}
    \label{fig:grid}
\end{figure}

To show that reference games are a suitable testbed for clarification behavior, we test three models on a controlled reference game dataset: the 7B and 72B versions of the Qwen2.5-VL \citep{Qwen25VL, QwenVL}, as its predecessor Qwen2-VL achieved the best performance on this task in \citet{junker-etal-2025-multimodal}, and GPT-5-mini \citep{gpt5-system-card}, as a representative commercial model. 

We use the color-grid reference game dataset \citep{mcdowell-goodman-2019-learning}, which consists of 197 games each with 60 rounds. In each round, a speaker sees three $3 \times 3$ color grids (one target and two distractors) and describes the target to a listener. The listener must then identify the target (see Figure~\ref{fig:grid} for an example). The dataset distinguishes three conditions with different levels of difficulty based on the color similarity between the target and the distractors: \emph{far} (easiest), \emph{split} (medium), and \emph{close} (hardest). These graded conditions make the task particularly well suited for quantifying how model behavior varies with referential difficulty. While the original setup allows multi-turn dialogue where dyads can interact freely, we focus on the initial speaker description only, to examine models' clarification behavior at the first turn.

While we evaluate the two Qwen2.5-VL models on the complete dataset, we limited GPT-5-mini's evaluation to a subset of 500 rounds due to API costs. For the Qwen2.5-VL models we report results both for the subset and for the full dataset.

\subsection{Experiments}
\label{sec:experiments}

\paragraph{Human Data}

Based on the English data from \citet{mcdowell-goodman-2019-learning}, we compute \textbf{human accuracy} on the color-grid reference game in terms of the ratio of successful trials. Overall human accuracy across all conditions (on the full data) is $0.92$; conditions individually increases from
    \emph{close} ($0.91$) to 
    \emph{split} ($0.93$) to 
    \emph{far} ($0.97$), 
see Table~\ref{tab:human-acc}.

\paragraph{Baseline Experiment}

Our baseline experiment is a conceptual replication of the experiment described in \citet{junker-etal-2025-multimodal}. We prompt the models with a concatenated image of the three color grids and the speaker's utterance (see Appendix~\ref{app:prompts} for the prompt). The model has to predict the position of the target, playing the listener role in the game.

In contrast to \citet{junker-etal-2025-multimodal}, we sample each model five times per round and use the majority vote as the predicted answer. We compute \textbf{Baseline Accuracy} from this predicted answer, following the same procedure as for the human data. Diversity sampling \citep{VashurinFadeeva2025} provides a more reliable estimate of a model's performance and enables us to quantify model uncertainty and to compute a simple \textbf{Baseline Confidence} measured as the proportion of samples matching the majority prediction. 
This approach yields discrete confidence levels $\{0.4, 0.6, 0.8, 1.0\}$, e.g., if 4 out of 5 samples predict the same grid, the confidence is $0.8$.
For Qwen-72B, we additionally report information-based uncertainty estimates based on the model’s probability distribution restricted to the three answer options (see Appendix~\ref{app:information-based-uncertainty}).

\paragraph{Clarification Experiment}

In a second experiment, we test whether the models are capable of asking for clarification when they are uncertain about the referent. We modified the prompt from the baseline experiment and explicitly instructed the model to ask a clarification question when it is uncertain (see Appendix~\ref{app:prompts} for the prompt). 

We sampled each model once and used the model's answer to evaluate its clarification behavior, yielding 
    (i) \textbf{CR-Rate} (the proportion of examples where a clarification question is generated),
    (ii) \textbf{Accuracy} (the proportion of non-clarification responses that are correct), 
    and (iii) \textbf{Relaxed} accuracy (the proportion of responses that are either correct or clarification requests).

\paragraph{Interaction Experiment}
\label{sec:interaction-experiment}

To investigate whether the model's clarifications are effective, i.e., seek information that would help resolve the reference, we conducted a follow-up interaction experiment in which a human (in-the-loop) answered the model's clarification request, giving the model another final chance to succeed.

\begin{figure}
    \centering
    \includegraphics[width=\columnwidth]{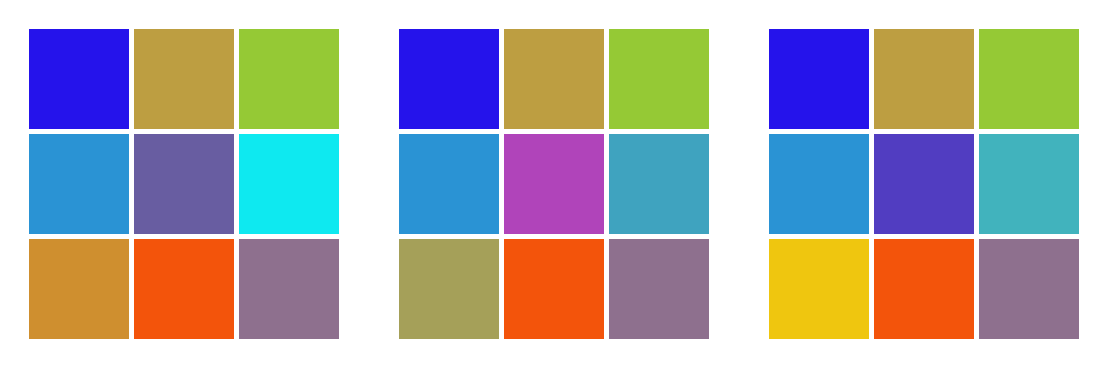}
    \begin{minipage}[t]{0.95\columnwidth}
    \footnotesize{
        \textsf{Human speaker:} 
            \emph{“Middle square is dark purple.”} (referring to the third item).\\
        \textcolor{gray}{\textsf{Qwen2.5-VL-72B Baseline:} 
            \emph{“First”} (incorrect).}\\
        \textsf{Qwen2.5-VL-72B Clarification:}
            \emph{“Could you please clarify which grid you are referring to?”}\\
        \textcolor{gray}{\textsf{Annotation}:
            not task-relevant (generic, fails to identify specific ambiguity)}\\
        \textsf{Human-in-the-loop}:
            \emph{“The grid with the darkest purple in the middle.”}\\
        \textsf{Qwen2.5-VL-72B}:
            \emph{“Third”} (correct).
    }
    \end{minipage}
    \caption{Example of Qwen2.5-VL-72B behavior with human-in-the-loop interaction.}
    \label{fig:grid22}
\end{figure}

We manually inspected all model-generated questions for Qwen-72B (116 items) and annotated whether each question was \emph{task-relevant} or \emph{not task-relevant}, based on the color grid and the original human description. For each clarification request, a human then responded to the model as a cooperative speaker. When the model’s clarification request was labeled task-relevant, the human answered it directly in order to resolve the ambiguity or underspecification that prompted the request. When the clarification request was not labeled task-relevant (see Figure~\ref{fig:grid22} for an example), the human instead reformulated the original utterance to provide a clearer or more detailed description of the target grid. Given the constrained nature of the reference game, we employed a single expert as annotator and human-in-the-loop (one of the authors). She was familiar with the color-grid dataset and experienced in provide clarification responses. To investigate the effects of clarification on model uncertainty and accuracy, we re-ran the Qwen-72B model on the clarified dialogues (prompt in Appendix~\ref{app:prompts}) and recomputed accuracy and confidence estimates, testing for their increase after clarification.

\section{Results}
\label{sec:results}

Table~\ref{tab:by_model_and_condition} shows the results for the baseline and clarification experiments across all models and conditions. Reported results are for the 500 item subset (19 items were excluded because GPT-5-mini produced null answers); for the two Qwen2.5-VL models, we additionally report performance on the full dataset for comparison, in parentheses. In the clarification experiment, it would be reasonable for the models to align their clarification behavior according to their uncertainty (as quantified in the baseline experiment): when confident about the target, they should respond with their prediction; when uncertain, they should generate a clarification request.\footnote{%
    Figure~\ref{fig:sankey} in the Appendix illustrates how model responses transition between the baseline and clarification experiments.
}

\begin{table*}[ht!]
\centering
\footnotesize
\begin{tabularx}{\textwidth}{lllllllll}
\toprule
    & & \multicolumn{2}{c}{\textbf{Baseline}} & \multicolumn{3}{c}{\textbf{Clarification}} & \multicolumn{2}{c}{\textbf{Comparison CR–Baseline}} \\
\cmidrule(lr){3-4}\cmidrule(lr){5-7}\cmidrule(lr){8-9}
    \textbf{Model} & \textbf{Cond.}
    & \textbf{Accuracy} & \textbf{Confidence}
    & \textbf{CR-Rate} & \textbf{Accuracy} & \textbf{Relaxed}
    & \textbf{Accuracy}& \textbf{Confidence} \\
\midrule 
    Qwen2.5 & close & 0.52 (0.47) & 0.87 (0.85) & 0.0 (0.002) & 0.46 (0.40) & 0.46 (0.40) & ---  (0.22) & ---  (0.73)\\
    VL-7B   & split & 0.52 (0.51) & 0.87 (0.87) & 0.0 (0.001) & 0.44 (0.41) & 0.44 (0.41) & ---  (0.25) & ---  (0.80)\\
            & far   & 0.56 (0.57) & 0.89 (0.88) & 0.0 (0.002) & 0.46 (0.44) & 0.46 (0.44) & ---  (0.14) & ---  (0.83)\\
            & ALL   & 0.53 (0.52) & 0.88 (0.87) & 0.0 (0.002) & 0.46 (0.42) & 0.46 (0.42) & ---  (0.20) & ---  (0.78)\\
\midrule
    Qwen2.5 & close & 0.68 (0.65) & 0.90 (0.90) & 0.24 (0.23) & 0.63 (0.64) & 0.71 (0.73) & 0.70 (0.57) & 0.86 (0.84) \\
    VL-72B  & split & 0.75 (0.71) & 0.91 (0.92) & 0.23 (0.24) & 0.73 (0.69) & 0.79 (0.76) & 0.76 (0.62) & 0.89 (0.87) \\
            & far   & 0.86 (0.78) & 0.93 (0.94) & 0.26 (0.26) & 0.85 (0.79) & 0.89 (0.85) & 0.86 (0.68) & 0.87 (0.89) \\
            & ALL   & 0.77 (0.71) & 0.91 (0.92) & 0.24 (0.24) & 0.73 (0.71) & 0.80 (0.78) & 0.78 (0.63) & 0.87 (0.87) \\
\midrule   
    GPT-5   & close & 0.87        & 0.97        & 0.17        & 0.91        & 0.92        & 0.58        & 0.94        \\
    mini    & split & 0.91        & 0.99        & 0.17        & 0.90        & 0.93        & 0.78        & 0.97        \\
            & far   & 0.98        & 1.00        & 0.06        & 0.98        & 0.99        & 0.89        & 1.00        \\
            & ALL   & 0.91        & 0.99        & 0.13        & 0.94        & 0.94        & 0.71        & 0.96        \\
    \bottomrule
\end{tabularx}
\caption{%
    Performance and clarification behavior by model and condition (close, split, far, and ALL combined). The table shows the results for the \textbf{Baseline} experiment, the \textbf{Clarification} experiment, and for the \textbf{Comparison} of clarification requests (CR) and baseline. Numbers reported are on the 500 item subset, and, for the two Qwen models, also on the full dataset (in parentheses).}
\label{tab:by_model_and_condition}
\end{table*}

In the \textbf{Baseline Experiment}, we evaluate how accurately and confidently models identify the target without any opportunity for clarification.
Model accuracy and confidence decrease with task difficulty (far > split > close), mirroring human performance (Table~\ref{tab:human-acc} in the Appendix) and previous findings \citep{junker-etal-2025-multimodal}.
GPT-5-mini reaches the highest baseline accuracy of 91\% and has a very high confidence of 99\% in its decisions, followed by Qwen-72B with 77\% (71\% on the subset) accuracy and slightly lower, but still high confidence of 91\% (92\%). Qwen-7B's baseline accuracy is considerably lower, 53\% (52\%), again with slightly lower but still high confidence of 88\% (87\%).
The high confidence across all models indicates a disposition to be overconfident, particularly for the Qwen.

In the \textbf{Clarification Experiment}, we examine how frequently models generate clarification requests and how clarification behavior interacts with performance and confidence.
GPT-5-mini generates clarification questions in 13\% of items, while Qwen-72B does so in 24\% (both for the subset as well as for the full dataset). Qwen-7B, in contrast, almost never generates clarification requests ($< 0.1\%$).
Clarification request rates also vary systematically across difficulty conditions for GPT-5-mini, rising from 6\% (far) to 17\% (split and close). This is not the case for Qwen-72B, where rates do not systematically track task difficulty.

For GPT-5-mini, accuracy of generated predictions (i.e., no clarification requests) exceeds its baseline accuracy (94\% vs. 91\%) consistent with reasonable behavior of only generating a response when confident. It even matches human performance (see Table~\ref{tab:human-acc}).
Qwen-72B's accuracy of generated predictions on the subset was slightly lower (73\% vs. 77\% baseline) with no difference on the full dataset (71\%). 
Under the relaxed accuracy measure, GPT-5-mini stays at 94\%, while Qwen-72B improves to 80\% (78\% on the full dataset). These increases suggest that clarification requests are generated for items that would otherwise result in wrong responses.

We now make a \textbf{Comparison} between clarification requests and baseline, examining how clarification behaviour relates to baseline accuracy and confidence. We begin by examining confidence patterns. Across models, items that elicited clarification requests have slightly lower confidence values than baseline confidence, which one could expect from a reasonable model that links clarification behavior to internal uncertainty.
However, confidence remains uniformly high, revealing overconfidence especially for the Qwen models. For instance, Qwen-72B's mean confidence only drops from 91\% to 87\% for items where it generated a clarification request.

Accuracy patterns show clearer differences when comparing clarification requests with the baseline.
For GPT-5-mini, items that elicited clarification requests have substantially lower accuracy than the baseline (71\% vs. 91\%), suggesting that the model tends to generate clarification requests on difficult items. This pattern is not stable for Qwen-72B. For the subset, accuracy is actually moderately higher than baseline (78\% vs. 77\%), but this reverses on the full dataset (63\% vs. 71\%), indicating that it tends to generate clarification requests on more difficult items. 
Qwen-7B, generating almost no clarification requests, in contrast, does not allow for meaningful comparison. On the few items of the full dataset where it generated one, accuracy is much lower than baseline (20\% vs. 52\%) though.

\begin{table*}[ht]
\footnotesize
\begin{tabularx}{\linewidth}{llXXXXXX}
\toprule
     & & \multicolumn{2}{c}{\textbf{Before}} 
     & \multicolumn{2}{c}{\textbf{After}} & \multicolumn{2}{c}{$\boldsymbol{\Delta}$} \\
\cmidrule(lr){3-4}\cmidrule(lr){5-6}\cmidrule(lr){7-8}
    \textbf{Items} &
    \textbf{Condition} &
    \textbf{Accuracy} &
    \textbf{Confidence} &
    \textbf{Accuracy} &
    \textbf{Confidence} &
    \textbf{Accuracy} &
    \textbf{Confidence} \\
\midrule
\textbf{CR-only} & close & 0.703 & 0.859 & 0.784 & 0.859 & $+$0.081 & $+$0.000 \\
    & split & 0.757 & 0.886 & 0.649 & 0.876 & $-$0.108 & $-$0.010 \\
    & far   & 0.857 & 0.867 & 0.786 & 0.962 & $-$0.071 & $+$0.095 \\
    & ALL   & 0.776 & 0.871 & 0.741 & 0.902 & $-$0.035 & $+$0.031 \\
\midrule
\textbf{full} & close & 0.682 & 0.898 & 0.701 & 0.898 & $+$0.019 & $+$0.000 \\
    & split & 0.753 & 0.912 & 0.728 & 0.910 & $-$0.025 & $-$0.002 \\
    & far   & 0.864 & 0.930 & 0.846 & 0.954 & $-$0.018 & $+$0.024 \\
    & ALL   & 0.767 & 0.914 & 0.759 & 0.921 & $-$0.008 & $+$0.007 \\
\bottomrule
\end{tabularx}
\caption{
    Accuracy and confidence of the Qwen2.5-VL-72B model in the \textbf{Interaction Experiment} by condition (close, split, far, and ALL combined).
    \textbf{Before} corresponds to the original speaker utterance-only baseline prompt, whereas \textbf{After} replaces the utterance with a dialogue-conditioned prompt containing the human-in-the-loop response. $\Delta$ is the difference between Before and After.
    The upper part of the table (CR-only) reports on the subset of 116 items where the model requested clarification. The lower part (full) reports on all items. Here, ‘after’ keeps the 365 items without clarification requests unchanged and replaces the 116 items on which the model previously asked a clarification request with the dialogue-conditioned prompts using human-provided clarification answers.
    We report majority-vote accuracy and mean consistency confidence.
}
\label{tab:interaction-experiment}
\end{table*}

In the \textbf{Interaction Experiment}, we investigate the quality of the model’s clarification requests and their impact on model confidence and accuracy. We find that only 42\% of the clarification questions generated by Qwen-72B are task-relevant. We further show that providing human answers to these clarification requests does not improve model performance (see Table~\ref{tab:interaction-experiment}). Although confidence increases after clarification, end-to-end accuracy decreases slightly (except in the close condition), suggesting either that the model does not effectively utilize the provided information, or that the clarification requests themselves are not useful, because they are not task-relevant or do not seek information the model actually requires.

\section{Discussion and Conclusion}
\label{sec:discussion-conclusion}

We examined whether vision-language models (VLMs), as listeners in reference games, can act on internal uncertainty by requesting clarification.

Our results show that VLMs display only a limited ability to do so. GPT-5-mini achieves high accuracy overall and uses clarifications more often when uncertain. In contrast, the Qwen2.5-VL models' use of clarifications is be largely decoupled from their internal uncertainty and overall task performance. Given the relative simplicity and contained nature of reference games, these findings highlight a gap in pragmatic abilities: even in a setting where uncertainty is quantifiable and clarification needs should be apparent, models struggle to translate uncertainty into appropriate interactional responses. 

This conclusion is reinforced by our interaction analysis that shows that providing human responses to these requests rarely improves performance. Many of the generated clarification requests are uninformative or even inappropriate given the task, rather than genuine attempts to resolve internal uncertainty -- let alone reflecting the universal principles for shared responsibility in achieving intersubjectivity \citep{dingemanse2015universal}.

\section*{Limitations}
\label{sec:limitations}

Our experiment has limitations. 
In the human data \citep{mcdowell-goodman-2019-learning}, dyads interacted for 60 rounds, allowing participants to build common ground \citep{Clark_WilkesGibbs_1986, Clark_1996} and develop conceptual pacts \citep{brennan1996conceptual}. In contrast, the VLMs lack access to this iterative grounding process, which may put them at a disadvantage compared to human participants. 
Additionally, the models' uniformly high confidence, based on diversity sampling, may indicate poor calibration. Information-based uncertainty quantification methods may generally offer a more accurate reflection of internal uncertainty. However, while for Qwen-72B, such estimates show lower but more graded confidence values, the model remains overconfident and does not better align clarification behavior with uncertainty (Appendix~\ref{app:information-based-uncertainty}).
Finally, the accuracy of the baseline models could stem from potentially flawed speaker descriptions (mistakes in their utterance, describing the wrong grid), rather than always from the listener.

\section*{Ethics Statement}

We do not consider this work to pose ethical concerns, and, as such, it did not require an ethics review. The dataset we use is publicly available.

\section*{Acknowledgments}

This research has been funded by the \href{https://www.dfg.de/}{Deutsche Forschungsgemeinschaft} (DFG, German Research Foundation) -- \href{https://gepris.dfg.de/gepris/projekt/512393437}{CRC-1646, project no. 512393437}, project \href{https://gepris.dfg.de/gepris/projekt/537416633}{B02}. We also acknowledge support from SAIL: “SustAInable Life-cycle of Intelligent Socio-Technical Systems” (Grant ID NW21-059A), an initiative of the Ministry of Culture and Science of the German state of North Rhine-Westphalia.

\bibliography{bibliography}

\clearpage
\appendix

\section{Implementation Details}
\label{app:implementation-details}

We used the following models for our experiment:
\begin{itemize}
    \itemsep0em
    \item \href{https://huggingface.co/Qwen/Qwen2.5-VL-7B-Instruct}{Qwen2.5-VL-7B-Instruct}: temperature 0.7
    \item \href{https://huggingface.co/Qwen/Qwen2.5-VL-72B-Instruct}{Qwen2.5-VL-72B-Instruct} (quantized): temperature 0.7
    \item \href{https://platform.openai.com/docs/models/gpt-5-mini}{GPT-5-mini}: No temperature control
\end{itemize}

We used a single NVIDIA RTX A6000 GPU for Qwen model inference. Depending on model size, generating responses took between \SI{4}{h} and \SI{96}{h} for the data. API costs for GPT-5-mini inference were approximately 5 USD.

\section{Scientific Artifacts}
\label{app:scientific-artifacts}

In our work, we mainly used scientific artifacts in the form of publicly available datasets and model implementations (MIT, Apache 2.0).
The color grid dataset is available on \href{https://github.com/forkunited/ltprg}{GitHub} (MIT License). We are confident that our work is consistent with their intended use.

\section{Information-Based Uncertainty Estimates}
\label{app:information-based-uncertainty}

In our experiment, we used consistency-based uncertainty estimations as it is the only uncertainty signal uniformly available across all models we evaluate, making it the fairest comparison. There are other, model-internal ways of quantifying uncertainty, e.g., using information theory, but these measures are not readily applicable when working with commercial models (such as GPT-5-mini).

In order to investigate, whether such information-based metrics would change our results, we explored them for the open-weights model Qwen-72B. Specifically, we re-computed confidence values using MSP (Maximum Softmax Probability) based on the model's probability distribution restricted to the three answer options.  Table~\ref{tab:msp-confidence-qwen72b} shows that, overall, the resulting confidence values are lower (compared to consistency-based confidence; Table~\ref{tab:by_model_and_condition}) and vary more gradually with task difficulty. Similar to the consistency-based measure, MSP confidence decreases on items where the model requests clarification, but this reduction is relatively small.
Despite the small improvement in confidence estimation, the model nonetheless remained overconfident when using information-based metrics.

\begin{table*}[t]
\footnotesize
\begin{tabularx}{\textwidth}{XXXXX}
\toprule
     & \multicolumn{2}{c}{\textbf{Baseline}} 
     & \multicolumn{2}{c}{\textbf{Comparison CR--Baseline}} \\
\cmidrule(lr){2-3}\cmidrule(lr){4-5}
    \textbf{Cond.}
    & \textbf{Accuracy} & \textbf{MSP Confidence}
    & \textbf{Accuracy} & \textbf{MSP Confidence} \\
\midrule
        close & 0.66 & 0.80 & 0.70 & 0.73 \\
        split & 0.75 & 0.83 & 0.68 & 0.77 \\
        far   & 0.88 & 0.88 & 0.88 & 0.80 \\
        ALL   & 0.76 & 0.84 & 0.76 & 0.77 \\
\bottomrule
\end{tabularx}
\caption{Qwen-72B performance and \emph{information-based confidence estimate} by condition (CLOSE, SPLIT, FAR, and ALL combined).
The table shows the results for the baseline experiment with MSP (maximum softmax probability) confidence results and the comparison of clarification requests (CR) and baseline. Numbers reported are on the 500-item subset. We use the maximum MSP as a confidence score and select the highest-probability option to predict the VLM answer.}
\label{tab:msp-confidence-qwen72b}
\end{table*}

\section{Prompts}
\label{app:prompts}

{\small\ttfamily\noindent
\begin{tabularx}{\linewidth}{X}
\toprule
    \textbf{Baseline Experiment Prompt} \\
\midrule
    You see three color grids arranged from left to right. \\
    A speaker describes one of these grids:\\ 
    \{utterance\} \\
    \\
    Which grid is the speaker referring to? Answer with exactly one word: first, second, or third. \\\\
\midrule
    \textbf{Clarification Experiment Prompt}\\
\midrule
    You see three color grids arranged from left to right. \\
    A speaker describes one of these grids: \{utterance\} \\
    \\
    INSTRUCTIONS:\\
    - If you can clearly identify which grid (first, second, or third) the speaker means, respond with exactly one word: first, second, or third.\\
    - If you are uncertain, unclear, or cannot confidently determine which grid is meant, you MUST ask a clarifying question starting with ''QUESTION:"\\
    \\
    Response:\\\\
\midrule
    \textbf{Interaction Experiment Prompt}\\
\midrule
    You see three color grids arranged from left to right. \\
    A speaker describes one of these grids: \{utterance\} \\
    The listener asked for clarification: \{question\}\\
    The speaker clarified: \{answer\}"\\
    \\

    Which grid is the speaker referring to? Answer with exactly one word: first, second, or third."""\\
    \\
    Response:\\\\
\bottomrule
\end{tabularx}}

\section{Additional Material}
\label{app:additional_material}

\begin{figure*}
    \centering
    \includegraphics[page=1]{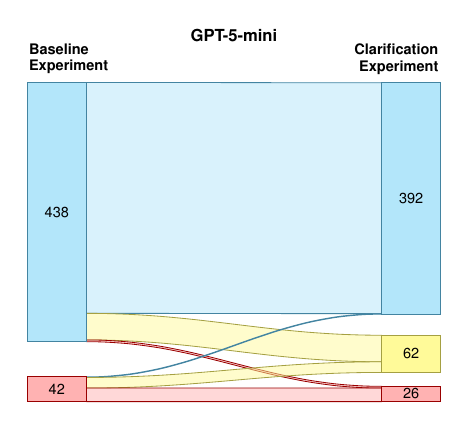}
    \includegraphics[page=2]{figures/sk-diagram.pdf}\\
    \vspace{-0.3cm}
    \includegraphics[page=3]{figures/sk-diagram.pdf}\vspace{-0.3cm}
    \caption{Sankey diagrams for GPT-5-mini (left) and Qwen2.5-VL-72B (right) showing how each model’s outcomes change from the baseline to the clarification experiment (on the 500-item subset). The flow indicates for which baseline items clarification requests were generated and how consistent correct and incorrect responses were.}
    \label{fig:sankey}
\end{figure*}

\begin{figure}
    \centering
    \includegraphics[width=\columnwidth]{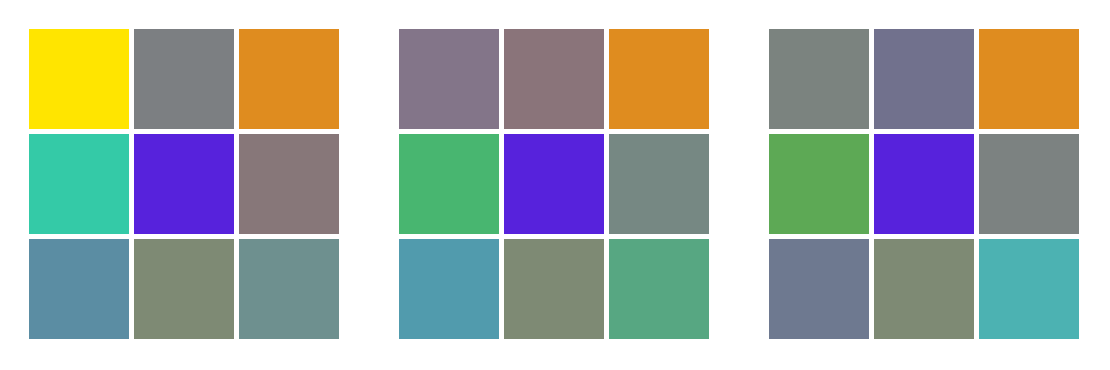}
    \begin{minipage}[t]{0.95\columnwidth}
    \footnotesize{
        \textsf{Human speaker:} \emph{“bottom right is grey not the bluish not the greenish”} (referring to the first item).\\
        \textsf{GPT-5-mini:} \emph{“QUESTION: Which of the three grids do you mean — the left, middle, or right one? None of their bottom-right squares looks plainly grey, so can you confirm which grid or clarify the shade you mean?”}
    }
    \end{minipage}
    \caption{Example of GPT-5-mini behavior in the Clarification Experiment.}
    \label{fig:grid1}
\end{figure}

Figure~\ref{fig:sankey} visualizes for which baseline items clarification requests were generated and how consistent responses were.
Figure~\ref{fig:grid1} shows an example of clarification requests generated by GPT-5-mini.
Table~\ref{tab:human-acc} reports the human performance on the color-grid data.

\begin{table}
\footnotesize
\setlength{\tabcolsep}{6pt}
\begin{tabularx}{\columnwidth}{l *{2}{>{\centering\arraybackslash}X}}
\toprule
\textbf{Condition} & \textbf{Subset} & \textbf{Full Dataset} \\
\midrule
    close  & 0.91 & 0.90 \\
    split  & 0.93 & 0.92 \\
    far    & 0.97 & 0.96 \\
\midrule
    ALL    & 0.94 & 0.92 \\
\bottomrule
\end{tabularx}
\caption{Human accuracy by condition on the subset and full dataset.}
\label{tab:human-acc}
\end{table}

\section{Use of AI Assistants}

AI assistants were used during manuscript preparation solely for specific linguistic reformulation to refine clarity and style, and to assist with code writing.
No AI-generated content was used to interpret results, and all conclusions were drawn by the authors.

\end{document}